\pdfoutput=1
\documentclass[11pt]{article}
\usepackage[final]{acl}

\usepackage{times}
\usepackage{latexsym}

\usepackage[T1]{fontenc}
\usepackage[utf8]{inputenc}

\usepackage{microtype}

\usepackage{inconsolata}

\usepackage{graphicx}

\usepackage{enumitem}
\usepackage{booktabs}
\usepackage{comment}

\usepackage{lscape} % For landscape mode if needed
\usepackage{longtable} % For tables spanning multiple pages
\usepackage{adjustbox} % For adjusting table width
\usepackage{arabtex} % Arabic text support
\usepackage{utf8} % UTF-8 support for Arabic
\setcode{utf8}
\usepackage{array} % for better table readability
\usepackage{amsmath}
\usepackage{multirow}

\title{Enhancing Arabic Automated Essay Scoring \\ with Synthetic Data and Error Injection}
\author{Chatrine Qwaider,\textsuperscript{1} Bashar Alhafni,\textsuperscript{1,2} Kirill Chirkunov,\textsuperscript{1} \\ \textbf{Nizar Habash,\textsuperscript{1,2}} \textbf{Ted Briscoe\textsuperscript{1}}\\
  \textsuperscript{1}MBZUAI, \textsuperscript{2}New York University Abu Dhabi\\
  \texttt{\{chatrine.qwaider,kirill.chirkunov,ted.briscoe\}@mbzuai.ac.ae}\\
  \texttt{\{alhafni,nizar.habash\}@nyu.edu}
  }

\begin{document}
\maketitle
\begin{abstract}
Automated Essay Scoring (AES) plays a crucial role in assessing language learners' writing quality, reducing grading workload, and providing real-time feedback. The lack of annotated essay datasets inhibits the development of Arabic AES systems. This paper leverages Large Language Models (LLMs) and Transformer models to generate synthetic Arabic essays for AES. We prompt an LLM to generate essays across the Common European Framework of Reference (CEFR) proficiency levels and introduce and compare two approaches to error injection. We create a dataset of 3,040 annotated essays with errors injected using our two methods. Additionally, we develop a BERT-based Arabic AES system calibrated to CEFR levels. Our experimental results demonstrate the effectiveness of our synthetic dataset in improving Arabic AES performance. We make our code and data publicly available.\footnote{\url{https://github.com/mbzuai-nlp/arabic-aes-bea25}}
\end{list} %known Arabtex bug
\end{abstract}

\section{Introduction}

Automated Essay Scoring (AES) is a technology that automates the evaluation and scoring of essays to assess language learners' writing quality while eliminating the need for human intervention \cite{shermis2003automated}. AES has gained great interest due to its significant benefits in the field of education \cite{lagakis2021automated,susanti2023automatic}. AES systems help teachers evaluate many essays with consistent scoring and reduced workload. On the other hand, AES helps students improve their writing quality through rapid real-time scoring and feedback \cite{hahn2021systematic}.

Unlike for English, it is difficult to develop robust and scalable AES systems for Modern Standard Arabic (MSA), primarily due to the lack of essay datasets necessary for building effective Arabic AES \cite{lim2021comprehensive, elhaddadi2024automatic}. This paper presents a framework to tackle the issue of data scarcity and quality by utilizing Transformers and Large Language Models (LLMs) to generate and build a synthetic dataset. 

Our approach begins with prompting GPT-4o to generate a variety of Arabic essays covering multiple topics and different writing proficiency levels as defined by the Common European Framework of Reference (CEFR) \cite{cefr2001}. Subsequently, we use a controlled error injection model to introduce errors into the correct Arabic essays, ensuring that erroneous essays reflect the type of errors that are commonly made by learners of Arabic in real-world scenarios. Our error injection approach consists of two steps: (i)~\textit{Error Type Prediction}, where a fine-tuned CAMeLBERT MSA model \cite{inoue-etal-2021-interplay} classifies the most likely error type for each word, and 
(ii)~\textit{Error Realization}, where we apply a bigram MLE model to determine the most probable transformation for each predicted error type. Our framework enables the generation of realistic human-like essays, enhancing data augmentation for Arabic AES systems.

Our main contributions are as follows: 
\vspace{-5pt} 
\begin{itemize}[noitemsep]
    \item Proposing a framework based on LLMs and Transformers for augmenting Arabic essays that accurately reflect human writing patterns.
    \item Creating a synthetic Arabic AES dataset with 3,040 essays annotated with CEFR proficiency levels.
    \item Developing an Arabic AES system using a BERT-based model, enabling accurate and scalable evaluation of Arabic essays based on CEFR standards.
\end{itemize}

The rest of the paper is organised as follows: \S\ref{related-work} reviews related work on AES, \S\ref{data} describes the dataset, and \S\ref{data-augmentation} outlines our data augmentation approach. \S\ref{error-injection} details the error injection methods, followed by an evaluation in \S\ref{evaluation}. We discuss our results in
\S\ref{discussion}
and \S\ref{conclusions}  presents the conclusion and future work.

%The rest of the paper is organized as follows: section 2 shows the related works in the Automated Essay scoring. The used data is described in section 4, while section 4 explains the data augmentation approach. In section 5, we describe two methods for error injections and evaluate the whole approach in section 6. Discussion and conclusion are in section 7, we address potential future works in section 8.

\section{Related Work}
\label{related-work}

AES has been investigated extensively, particularly in English \cite{lim2021comprehensive, ramesh2022automated}, where multiple tools have been introduced such as IntelliMetric \cite{elliott2003overview}, e-rater \cite{attali2006automated}, Grammarly,\footnote{\url{https://app.grammarly.com/}} Write and Improve\footnote{\url{https://writeandimprove.com/}} \cite{yannakoudakisetal}, and others. The development of English AES systems has been enabled by large scale annotated datasets such as the First Cambridge English (FCE) dataset\footnote{\url{https://ilexir.co.uk/datasets/index.html}} \cite{yannakoudakis}, Automated Student Assessment Prize (ASAP) dataset,\footnote{\url{https://www.kaggle.com/c/asap-aes}} the TOEFL11 corpus \cite{blanchard2013toefl11}, and the ICLE (International Corpus of Learner English) \cite{granger2003international}. These datasets contain thousands of student essays with proficiency level grades, often along multiple dimensions. 

In contrast, Arabic AES research has received less attention. Some studies have applied feature engineering and machine learning to develop models~\cite{alghamdi2014hybrid,al2016automated,alobed2021adaptive,gaheen2021automated}, but they partially address key challenges, especially the scarcity of large, publicly available annotated datasets for improving Arabic writing quality.
%This work and related research has only partially addressed the key challenges hindering the development of Arabic AES models, namely, the lack of publicly available large-scale annotated datasets designed to improve the quality of Arabic writing. 

%Ghazawi and Simpson 
%\newcite{ghazawi2024automated} introduced AR-AES, a benchmark dataset that contains 2,046 essays written by undergraduate students across three different university faculties, featuring a mix of question types. Two educators from each faculty  annotated and graded the essays using clearly defined rubrics, aiming to assess students' academic performance across various disciplines. Unlike AR-AES, our work aims to evaluate and assess writing proficiency using the CEFR rubrics as these constitute an internationally accepted standard.
\newcite{ghazawi2024automated} introduced AR-AES, a benchmark of 2,046 undergraduate essays from three university faculties, annotated by two educators per faculty using rubrics to assess academic performance. In contrast, our work focuses on writing proficiency, using the CEFR standard.

\newcite{bashendy2024qaes} %is considered a significant contribution, that 
presented QAES, the first publicly available trait specific annotations for Arabic AES.  QAES extends the Qatari Corpus of Argumentative Writing (QCAW) \cite{ahmed2024building}, which consists of 195 Arabic argumentative essays. They implemented multi-layered annotation of traits such as coherence, organization, grammar, and others. Despite its comprehensive annotation, it is small in size and limited to two prompts. While QAES multi-traits scores are publicly available, the QCAW holistic score is not.

\newcite{habash-palfreyman-2022-zaebuc} presented the Zayed University Arabic-English Bilingual Undergraduate Corpus (ZAEBUC).
%, a bilingual non-parallel Arabic-English corpus.
This corpus comprises non-parallel essays in Arabic and English related to three prompts collected from first-year university students with differing writing proficiency. ZAEBUC includes 216 annotated Arabic essays featuring manual annotations for syntactic and morphological characteristics and a CEFR-based proficiency assessment. Again, ZAEBUC is small in size and limited to three prompts.

%Researchers have investigated several approaches to data augmentation such as sampling techniques, introducing noise into texts, or paraphrasing, to overcome the challenges of data scarcity and quality \cite{li2022data}. 
Researchers have explored data augmentation methods like sampling, noise injection, and paraphrasing to address data scarcity and quality~\cite{li2022data}. The recent development of LLMs has paved the way for researchers to explore promising new data synthesis solutions \cite{wang2024survey, long2024llms}. Transformers and LLMs can closely mirror real-world distributions while introducing valuable variations across multiple tasks and domains \cite{wang2024autosurvey}. 

%GPT models such as ChatGPT, GPT-4, and GPT-4o demonstrate significant capabilities in augmenting and generating synthetic essays for the purpose of English AES \cite{ramesh2022automated}.  LLMs and Transformer models with Arabic capabilities have shown robust generalization across different Arabic NLP tasks such as: Question-Answering \cite{samuel-etal-2024-llms}, Code-Switching \cite{alharbi2024leveraging}, Name Entity Recognition (NER) \cite{sabty2021data}, Grammar Error Correction (GEC) \cite{solyman2023optimizing}, and Sentiment Analysis \cite{refai2023data}. 
GPT models %like ChatGPT, GPT-4, and GPT-4o 
have shown strong capabilities in generating synthetic essays for English AES~\cite{ramesh2022automated}. LLMs and Transformers have also generalized well in Arabic NLP tasks, including Question Answering~\cite{samuel-etal-2024-llms}, Code Switching~\cite{alharbi2024leveraging}, NER~\cite{sabty2021data}, Grammatical Error Correction~\cite{alhafni2025enhancingtexteditinggrammatical,solyman2023optimizing}, and Sentiment Analysis~\cite{refai2023data}. 
However, to the best of our knowledge no research has utilized such models to generate Arabic essays  across CEFR writing proficiency levels.

%Unlike previous works that rely on manual annotation or traditional augmentation techniques (e.g., back-translation), our work leverages GPT to generate human-like Arabic essays tailored to CEFR standards. Additionally, we fine-tune AES scoring using BERT-based models, ensuring consistency between real and synthetic essays to optimize proficiency-level predictions.

%Mention existing work on GPT or LLM-based essay scoring for English (e.g., OpenAI's work, research on using GPT-3/4 for scoring). Comparison with Previous Data Augmentation Techniques. Many English AES studies have used synthetic essay generation, back-translation, and augmentation strategies. How does your approach with GPT differ or improve upon these methods?

%\newpage

%\input{Tables/zprompt}

\section{Data: The ZAEBUC Corpus}
\label{data}
For all our experiments, we use the 
ZAEBUC corpus \cite{habash-palfreyman-2022-zaebuc}. ZAEBUC  comprises essays written by native Arabic speakers, which were manually corrected and annotated for writing proficiency using the  CEFR \cite{cefr2001} rubrics and scale.  Each essay was annotated by three CEFR-proficient bilingual speakers. \newcite{habash-palfreyman-2022-zaebuc}, assigned a holistic CEFR level to each essay by converting the three CEFR ratings into numerical scores (ranging from 1 to 6) and then taking the rounded average.
The essays in the corpus were limited to three  prompt choices on \textit{Social Media}, \textit{Tolerance}, and \textit{Development}; see Table~\ref{zprompts}. We use the splits created by \newcite{alhafni-etal-2023-advancements}. Table~\ref{tab:cefr-stats} shows the CEFR level distribution of the ZAEBUC corpus based on holistic CEFR scores.
%The ZAEBUC corpus presents an inherent limitation, as it is relatively small and exhibits a skewed distribution, with the majority of essays concentrated at the B1 and B2 levels, and no representation of A1 or C2 levels. This imbalance, common in available Arabic learner writing datasets, strongly motivated our synthetic data generation approach, which aims to address these gaps and achieve a more balanced CEFR-level distribution.
The ZAEBUC corpus is limited in size and skewed toward B1–B2 levels, with no A1 or C2 essays. This common imbalance in Arabic learner data motivated our synthetic approach to create a more balanced CEFR distribution.

\begin{table}[t]
\centering
 \setlength{\tabcolsep}{2pt}
 \begin{small}
\begin{tabular}{p{3in}}
\toprule
\multicolumn{1}{r}{ \<وسائل التواصل الاجتماعي وتأثيرها على الفرد والمجتمع.> }\\
 How do social media affect individuals and society?\\\midrule
\multicolumn{1}{r}{ \<كيف نعزز ثقافة التسامح في المجتمع؟> }\\
How can the UAE promote a culture of tolerance in society? \\ \midrule

\multicolumn{1}{r}{ \<التطور الحضاري الذي تشهده دولة الإمارات العربية المتحدة> }\\
What do you think are the most important developments in the UAE at the moment? \\\bottomrule
\end{tabular}
\end{small}
\caption{The prompts given to the essay writers in the ZAEBUC corpus \cite{habash-palfreyman-2022-zaebuc}.}
\label{zprompts} 
\end{table}
\begin{table}[!t]
\centering
\setlength{\tabcolsep}{6pt} % Adjust column spacing
\begin{tabular}{c c c}
\toprule
\textbf{CEFR Level} & \textbf{Count} & \textbf{Percentage} \\
\midrule
\textbf{A1} & 0  & 0\% \\
\textbf{A2} & 7  & 3\% \\
\textbf{B1} & 110 & 51\% \\
\textbf{B2} & 80  & 37\% \\
\textbf{C1} & 11  & 5\% \\
\textbf{C2} & 0   & 0\% \\
\textbf{Unassessable} & 6 & 3\% \\
\midrule
\textbf{Total} & 214 & 100\% \\
\bottomrule
\end{tabular}
\caption{ZAEBUC corpus CEFR level distributions.}
\label{tab:cefr-stats}
\end{table}

\section{Synthetic Data Augmentation}
\label{data-augmentation}
%data augmentation, which enhances existing data samples through transformations, and data synthesis, which creates entirely new samples from scratch or based on generative models
%chatrine: would like to discuss the difference, can we use the terms exchangeable
%Specifically, synthetic data is designed to mimic the characteristics and patterns of real-world data (Best practices and lessons learned on synthetic data for language models)

We propose a synthetic data augmentation approach leveraging the ZAEBUC dataset to generate synthetic essays that align with CEFR rubrics 
%the Common European Framework of Reference for Languages (CEFR) 
and have features similar to human text.
The pipeline utilizes three phases: Building Essay Prompts, Feature Profiling, and finally Data Augmentation.

\subsection{Building Essay Prompts} 
%We began by compiling a collection of essay prompts covering diverse categories and proficiency levels.
%While not taken directly from existing frameworks, The prompts draw inspiration from common themes in language assessments, such as placement tests and academic writing tasks. Our aim was to cover familiar, relevant, and level-appropriate topics—like social issues, education, and personal experiences—while ensuring balanced coverage across CEFR levels.

%We considered three proficiency levels aligned with the CEFR scale: Beginner (A1–A2), Intermediate (B1–B2), and Advanced (C1–C2). While some categories, such as Hobbies, are suitable for all levels, others, like Politics, Technology, and Education, were introduced specifically for advanced learners.
%Using various LLMs, including GPT-4o, Gemini, and Copilot, we generated 100 prompts across these categories, ensuring relevance for Arabic essay writing and an unbiased distribution across proficiency levels. We then manually reviewed the prompts, removing redundancies and irrelevant entries while maintaining comprehensive coverage.
%
We began by compiling a diverse set of essay prompts across various categories and CEFR levels. While not directly drawn from established frameworks, our prompts were inspired by themes common in language assessments, including placement tests and academic writing. We aimed to cover familiar and level-appropriate topics, such as social issues, education, and personal experiences, while ensuring balance across the CEFR bands. 
We considered three proficiency levels: Beginner (A1–A2), Intermediate (B1–B2), and Advanced (C1–C2). General themes, such as hobbies, suited all levels, while more complex topics, including politics, Technology, and Education, were reserved for advanced learners.

Using LLMs like GPT-4o\footnote{\url{https://openai.com/index/hello-gpt-4o/}}, Gemini\footnote{\url{https://gemini.google.com/app}}, and Copilot\footnote{\url{https://copilot.microsoft.com/}}, we generated 100 prompts, followed by a manual review to remove redundancies and ensure both relevance for Arabic essay writing and balanced proficiency coverage.
The final collection consists of 152 balanced and diverse prompts. Table \ref{tab:prompts_count} presents the selected categories and the distribution of the prompts across levels, while Table~\ref{tab:topic_prompts} provides example prompts for the Hobbies category.

%To generate Arabic essays and construct the dataset, we first need to build an essay prompts dataset. We started by setting the main categories we wanted to cover. Some categories, such as Hobbies, were suitable for all three proficiency levels (Beginner, Intermediate, and Advanced), while others, like Politics, Technology, and Education, were explicitly introduced to advanced learners. 
%Next, we utilized various LLMs, such as GPT-40, Gemini, and Co-pilot, to generate 100 prompts within these categories, ensuring they were relevant for Arabic writing essays and covered different student proficiency levels.  
%After getting the prompts, we carefully checked them; we eliminated all redundant and irrelevant prompts while ensuring comprehensive coverage across all the categories and levels. The final dataset was designed to be balanced and diverse, with 152 prompts/questions. The prompts are structured to reflect different proficiency levels as follows: (i) Beginner (A1–A2), (ii) Intermediate (B1–B2), (iii) Advanced (C1–C2).
%Table \ref{tab:prompts_count}  shows the defined categories and the distribution of the prompts across the levels, while Table \ref{tab:topic_prompts} shows some prompts example from the topic of Hobbies.

\begin{table}[t!]
    \centering
    \begin{tabular}{lccc}
        \toprule
        \textbf{Topic} & \textbf{B} & \textbf{I} & \textbf{A} \\\midrule
        %\hline
        Culture and Traditions & 1 & 3 & 2 \\ %\hline
        Daily Life & 2 & 2 & 2 \\ 
        Education & 3 & 6 & 8 \\ 
        Environment & 2 & 2 & 3 \\ 
        Future & 1 & 2 & 2 \\ 
        History and Culture & 2 & 2 & 2 \\ 
        Hobbies & 3 & 2 & 2 \\ 
        Imaginary & 5 & 2 & 2 \\
        Life/Time Management & 4 & 4 & 2 \\
        Personal Experiences & 7 & 2 & 2 \\ 
        Relations & 4 & 2 & 2 \\ 
        School Life & 4 & 2 & 2 \\ 
        Sport and Health & 2 & 3 & 1 \\ 
        Technology and Media & 2 & 8 & 6 \\ 
        Travel and Experience & 1 & 2 & 1 \\ 
        Politics and Government & 2 & 2 & 7 \\ 
        Social Issues & 2 & 7 & 6 \\ \midrule
        \textbf{Total} & 47& 53& 52 \\ 
        \bottomrule
    \end{tabular}
    \caption{Count of Arabic text prompts by level and topic. B: Beginner level (A1,~A2), I: Intermediate level (B1,~B2), A: Advanced level (C1,~C2).}
    \label{tab:prompts_count}
\end{table}

\begin{table*}[ht]
\tabcolsep5pt
\resizebox{\linewidth}{!}{
\begin{tabular}{l p{2.7in} p{3.7in} }
\toprule
\textbf{Level}    & \multicolumn{1}{c}{\textbf{Arabic Prompt}}& \multicolumn{1}{c}{\textbf{English Prompt}} \\ \midrule
\textbf{Beginner} &\multicolumn{1}{r}{\small \<ما هي هوايتك المفضلة؟ >}$\bullet$ & $\bullet$ What is your favorite hobby?            \\ 
  & \multicolumn{1}{r}{\small \<تحدث عن نشاط تحبه في عطلة نهاية الأسبوع.>}$\bullet$ & $\bullet$ Talk about a weekend activity you love. \\ 
  & \multicolumn{1}{r}{\small \< ما هي الرياضة التي تحب ممارستها؟ لماذا؟>}$\bullet$ & $\bullet$ What is your favorite sport? Why?       \\ \midrule
\textbf{Intermediate} &
  \multicolumn{1}{r}{\small \<في ممارستها؟ لماذا تحبها؟>
  \<ما هي هوايتك المفضلة وكيف بدأت >} $\bullet$ &
   $\bullet$ What is your favorite hobby and how did you start practicing it?  Why do you enjoy it?  \\ 
 &
 \multicolumn{1}{r}{\small  
 \< ما هي ولماذا تهمك؟>  
 \< هل ترغب في تعلم هواية جديدة؟> 
  } $\bullet$ &
   $\bullet$ Do you wish to learn a new hobby?  What is it and why does it interest you? \\ \midrule
   
\textbf{Advanced} &
 \multicolumn{1}{r}{\small  
  \<ناقش كيف تؤثر الهواية على صحتك النفسية والجسدية.>} $\bullet$ &
   $\bullet$ Discuss how hobbies impact your mental and physical health. \\
   
 &  \multicolumn{1}{r}{\small \<كيف يمكن أن تكون الهوايات وسيلة للتعبير عن الذات؟> }&
  How can hobbies serve as a means of self-expression?\\ 
  &
   \multicolumn{1}{r}{\small   \< هل هناك هواية جديدة ترغب في تجربتها؟ >} $\bullet$ & $\bullet$ Is there a new hobby you wish to try? Discuss the reasons  \\
&  \multicolumn{1}{r}{\small   \<ناقش الأسباب التي تجعلك مهتمًا بها وكيف تعتقد أنها ستفيدك.> }&
    you are interested in it and how you believe it will benefit you. \\ 
  \bottomrule
\end{tabular}
}
\caption{Examples of prompts related to the topic of Hobbies and classified into one of three different levels.}% (Beginner, Intermediate, Advanced) from the topic prompt dataset.}
\label{tab:topic_prompts}
\end{table*}

\subsection{Feature Profiling}
We construct linguistic profiles for each CEFR level using the ZAEBUC corpus. Each profile contains various levels of linguistic information.  
Representing different
lexical and syntactic features, we use the number of words/sentences (\(N_w\),\(N_s\)),  the  number of tokens/vocabulary (\(N_v\)),  words/sentences lengths (\(L_w\),\(L_s\)), and  sentence complexity measured by syntactic tree depth (\(D_s\)).

We define the lexical diversity (Type-Token Ratio, TTR) as:
\begin{equation}
    \text{TTR} = \frac{\text{Unique Tokens}}{\text{Total Tokens}}
\end{equation}
Similarly, we calculate the sentence complexity by:
\begin{equation}
    C_s = \frac{\sum_{i=1}^{N} D_i}{N}
\end{equation}
where \( D_i \) is the syntactic depth of sentence \( i \) and \( N \) is the total number of sentences.

For morphological features, we use the ZAEBUC morphological annotations:  the most frequent POS tags, such as nouns, verbs, adjectives, etc. 

We aggregate all extracted features  across the essays to get a quantitative representation at different writing CEFR levels, which serves as a reference for later stages.
%%Chatrine: shall we list all te features?

\subsection{Zero-shot Data Augmentation} 

%With the emergence of LLMs, the need for good prompt engineering has increased since the output of the LLM differs based on the prompts, the provided instructions, and the prompt language %chatrine: add citation. 
%Previous works in Arabic NLP reported that using English as an instruction language for the input prompts improved the outputs.%citation 
%In our approach, we exploit various prompts for a zero shot data augmentation to check which one can generate human-like text as well as follow the guidelines instructions. We exploit GPT-4o as the generated model since it is affordable and gives us more token space both for input and output. 
%
Effective LLM prompt engineering has become increasingly important, as the model's output varies based on the prompt, provided instructions, and prompt language.
Previous studies in Arabic NLP have shown that using English as the instruction language for input prompts can improve output quality \cite{kmainasi2024native,koto-etal-2024-arabicmmlu}. 

In our approach, we experiment with various prompts for zero-shot data augmentation to identify those that produce human-like text while adhering to guideline instructions. We use GPT-4o as our generation model due to its affordability and larger token capacity for both input and output.
%
%The GPT prompts contain
%
%\begin{itemize}[noitemsep]%\itemsep0em 
%    \item The target CEFR level
%    \item CEFR guidelines and instructions
%    \item The linguistic profile fot the targeted CEFR to control the prompts output 
%    \item The topic prompts/question from the aforementioned topic prompts dataset
%\end{itemize}
The GPT prompts include \textbf{(a)} the target CEFR level, \textbf{(b)} CEFR guidelines and instructions, \textbf{(c)} the linguistic profile for the targeted CEFR level to control the prompt output, and \textbf{(d)} the topic prompt or question from the previously mentioned topic prompts dataset.
For these missing levels (A1 and C2), instead of injecting a pre-defined profile, GPT-4o was directly prompted to act as an assistant and generate data based on the general standards and rubrics of the CEFR.

To check the quality of the generated essays and whether they follow the prompt instructions, 
we build a linguistic feature profile (vector) for each augmented essay. 
%We employ the cosine similarity  to assess the faithfulness of the augmented approach and whether the augmented  essays are close to the original CEFR profiles. 
 We then assess the alignment between the generated essays and the reference CEFR-level profiles by computing their feature vectors' cosine similarity as in equation \ref{eq:cosine}. %We applied the cosine similarity as in equation \ref{eq:cosine}  either to one of the features profiles (lexcial/syntactical features , morphological features or both of them). 
 Specifically, given two real-valued feature vectors 
\( P_i \) (the CEFR reference profile) and 
\( Q_i \)(the generated essay), the cosine similarity is calculated as:

\begin{equation}
    \cos(\theta) = \frac{\sum_{i} P_i Q_i}{\sqrt{\sum_{i} P_i^2} \cdot \sqrt{\sum_{i} Q_i^2}}
    \label{eq:cosine}
\end{equation}

%where \( P_i \) and \( Q_i \) represent the components of the linguistic feature vectors  in the original and generated distribution profiles. 
This metric ensures that the synthetic data closely aligns with real human essay patterns. Based on the computed similarity score, we assign a predicted CEFR level to each essay.

Later, we calculate the alignment between the predicted CEFR level and the target level specified in the GPT-4o prompt (ground truth) using the following agreement formula:

\begin{equation}
    \text{Agreement} = \frac{\sum_{i=1}^{n}  (\hat{y}_i = y_i)}{n}
\end{equation}
where \( \hat{y}_i \) is the predicted level and \( y_i \) is the ground truth. This process evaluates how well GPT-4o succeeded in aligning the generated content with the intended proficiency level, serving as a measure of agreement rather than a prediction from an external model.

%We conducted various prompt engineering refinement rounds to enhance the generated Arabic essays while ensuring they aligned with the requested CEFR level. 

We conducted multiple rounds of prompt engineering refinements to improve the quality of the generated Arabic essays and ensure alignment with CEFR levels.

First, we found that straightforward prompts without explicit controlled linguistic instructions and explanations resulted in incoherent essays, including irrelevant topics and English text, achieving only 20.5\% matching agreement with linguistic feature profiles. 
In a subsequent round, we introduced detailed definitions of linguistic features and restricted outputs to Arabic-only text, which improved agreement to 26\%. However, the model still occasionally produced incomplete essays and injected text from the prompt into the essay. 

%
%In the first round, we tested straightforward prompts that  included the aforementioned parameters. However, prompts lacked explicit instructions for interpretation of the features or how the LLM should generate a coherent and relevant Arabic essay. This resulted in problematic output  such as English essays or topically-irrelevant content
%We noticed that the model sometimes produced English essays, while the content of the Arabic essays was often irrelevant, that 
%resulted on 20.5\% matching agreement. 
%These observations led us to the next experiment, in which the prompt needed to provide clear guidance.
%

%In a subsequent round, we improved prompts by adding explicit definitions for the linguistic feature profiles, and  updated the instructions to generate only Arabic essays. We achieved an agreement of 26\%. However, the model still occasionally produced incomplete essays and  injected text from the prompt into the essay. % These findings indicate that while the alignment matching improved, however the prompt itself required further refinements to prevent prompt leakage into the augmented essays.
%

\begin{figure}[t]
  \fbox{\includegraphics[width=0.96\linewidth]{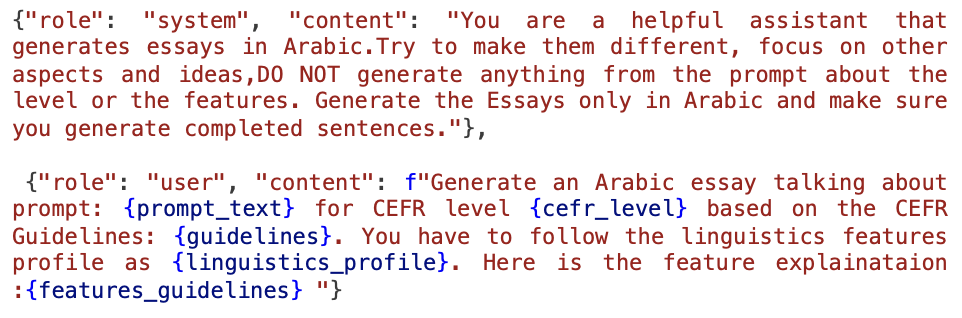}}
  \caption{GPT-4o prompts messages that have been used to generate Arabic essays}
  \label{fig:gpt_prompt}
\end{figure}

The most effective prompt structure format is illustrated in Figure \ref{fig:gpt_prompt}.  We  separated system-level control instructions from user-defined parameters, thereby providing clearer guidance for structured and proficiency-aligned text generation. This refinement increased agreement to 27.5\%, demonstrating that precisely controlled instructions enhance LLM performance in structured writing tasks.

%The last round of refinements focused on enhancing the control and instructional aspects of the prompts. Figure \ref{fig:gpt_prompt} illustrates the final structured prompt format, distinguishing two separate roles. In the system role, we include instructions to control the generation process, whereas in the user role, we input all parameters and guidelines to align with specific CEFR levels. Combining system role instructions with user-defined tuning parameters facilitated an improved data augmentation process, achieving a match agreement of 27.50\%. 
Ultimately, we generated 3,040 Arabic essays covering all CEFR levels and various topics, where each prompt was used to create ten essays. This effort was intentionally designed to address the imbalanced CEFR distribution in the original ZAEBUC corpus, where B-level essays were overrepresented. By constructing a more balanced synthetic dataset, we aimed to enhance model performance across the full proficiency spectrum. The structured and controlled prompt design also improved alignment with learner writing styles while providing a consistent framework for generating realistic Arabic essays. Table \ref{tab:cefr-gpt} presents the distribution of generated essays across different CEFR levels. The full dataset statistics are provided in Appendix~\ref{secb:appendix}.

%We ultimately generated  3,040 Arabic essays covering all topics in the prompt dataset, generating ten essays for each prompt. Table \ref{tab:cefr-gpt} shows the distribution of  generated essays across the CEFR levels.

\begin{table}[!t]
\centering
\setlength{\tabcolsep}{6pt} % Adjust column spacing
\begin{tabular}{c c c}
\toprule
\textbf{CEFR Level} & \textbf{Count} & \textbf{Percentage} \\
\midrule
\textbf{A1} & 470  & 15.5\% \\
\textbf{A2} & 470  & 15.5\% \\
\textbf{B1} & 530  & 17.4\% \\
\textbf{B2} & 530  & 17.4\% \\
\textbf{C1} & 520  & 17.1\% \\
\textbf{C2} & 520  & 17.1\% \\
\midrule
\textbf{Total} & 3,040 & 100\% \\
\bottomrule
\end{tabular}
\caption{The generated corpus CEFR level distributions.}
\label{tab:cefr-gpt}
\end{table}

\section{Error Injection}
\label{error-injection}
Human-generated text naturally contains some grammatical errors and linguistic infelicities.
%not all writers and Arabic learners are professionals in the language. 
%
In order to create human-like essays,  we need to add similar kinds of errors to the synthetic essays that reflect the level of writing attainment. In this phase, we prompt GPT-4o to inject errors into the previously generated essays while maintaining their aligned CEFR levels by utilizing error profiling. 
%We start with two initial steps to ensure a smooth and controllable error injection process.
%Next we discuss error profiling and then describe two different methods we built for error injection.

\subsection{Error Profiling}

\paragraph{Error Distribution Profiles}
To model the distribution of errors to inject  into the synthetic essays, 
we again leverage the ZAEBUC corpus, which contains the erroneous essays aligned with the manually corrected ones. 
We followed the same methodology we used to construct the linguistic feature profiles for each CEFR level to develop error distribution profiles aligned with CEFR levels. The error profile captures and reflects the authentic distribution patterns observed in human writing at different CEFR levels.

\begin{figure}[t]
  \fbox{\includegraphics[width=\columnwidth]{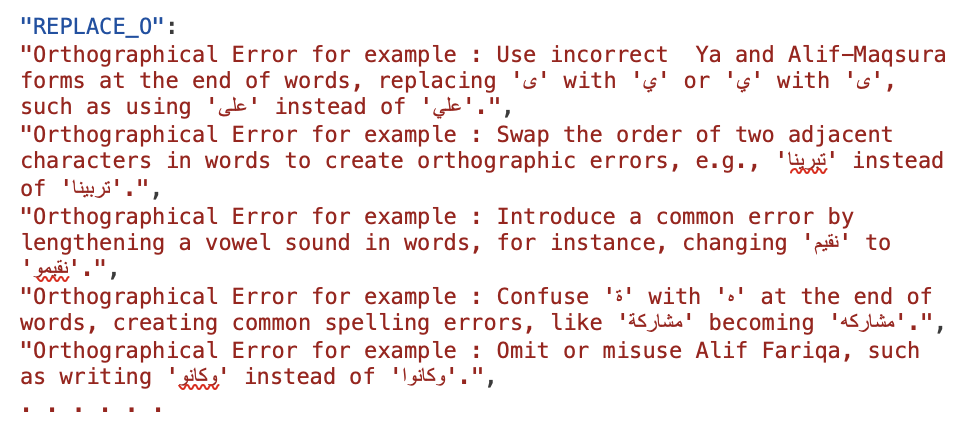}}
  \caption{An example of  orthographical error instructions from the developed errors instructions repository}
  \label{fig:error_instruction}
\end{figure}

\textbf{Developing an Error Instruction Repository}
To prompt GPT-4o to generate essays containing errors we applied the Grammatical Error Detection (GED) model proposed by \cite{alhafni-etal-2023-advancements} to the ZAEBUC corpus to annotate errors using 13 error tags and to obtain error distributions for each CEFR level. We created the repository using the error tags, where we also added a formal definition of what those tags describe in terms of linguistic errors. In addition, we expanded the error taxonomy by splitting it into finer-grained classes. Each error instruction is followed by an example showing the correct word and the erroneous version. The explanation was based on the \textit{extended ALC taxonomy} \cite{alfaifi2013error}, which was refined later and introduced as ARETA \cite{belkebir-habash-2021-automatic}. Appendix \ref{sec:error_taxonomy} presents examples of the error types. Figure~\ref{fig:error_instruction} shows some examples from our error instruction repository.

\subsection{GPT-Based Error Injection}
We prompted GPT-4o to inject errors into the synthetically generated essays based on the error distribution profiles while maintaining the CEFR level. The model processed one essay at a time in a zero-shot setting, except that we included the definition and explanation of the error tags. For example, \textbf{M} indicates a morphological error, while \textbf{Merge} targets two mistakenly split tokens that need to be merged, and so on. 

After conducting multiple experiments, we observed  the following issues: (i) The model struggled to follow the predefined error distribution perhaps due to the complexity of the prompts. (ii) The model was confused by certain error tags, particularly \textbf{Split} and \textbf{Merge}. %as well as the UNK (unknown errors). 
These errors were mainly ignored in the injected text. (iii) We calculated the cosine similarity between the main error profile and the injected essays' error distributions as shown in Equation~\ref{eq:cosine}. When we injected all errors at once, the similarity agreements did not exceed~20\%; however, when we reduced the number of error tags per essay %or omitted the misleading tags, 
the agreements significantly improved, reaching~86\%. %86.12\%.

\begin{figure}[t]
\fbox{\includegraphics[width=0.95\columnwidth]{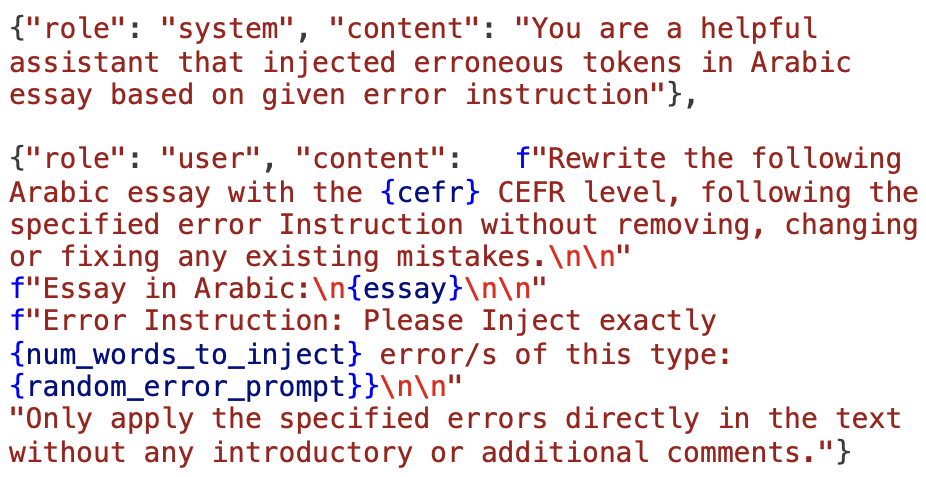}}  
  \caption{Sample GPT-4o error injection prompt}
  \label{fig:error_prompt}
\end{figure}

Therefore we implemented a method where each error type was injected separately. This required multiple iterations over the same essay, corresponding to the number of error tags shown in the error distribution profile for each CEFR level. Figure \ref{fig:error_prompt} shows an example of a GPT-4o prompt for error injection. Some error types, especially orthographic errors, are more frequent among Arabic writers than others.
The prompt was intentionally designed through prompt engineering. The `helpful assistant' component establishes a cooperative persona for the LLM, while the subsequent instruction to `inject erroneous tokens' explicitly guide GPT-4o towards the specific task of error introduction. This approach ensures that GPT-4o is not making random edits but is rather following predefined instructions to create targeted errors, aligning with the overall goal of generating realistic synthetic data.

To reflect this, we randomly select weighted error instructions based on the average frequency of each error type. Figure \ref{fig:error_strategy} shows the pseudocode for the selection process. The full pseudocode is in Appendix~\ref{sec:appendix}.

\begin{figure}[t]
\fbox{\includegraphics[width=0.95\columnwidth]{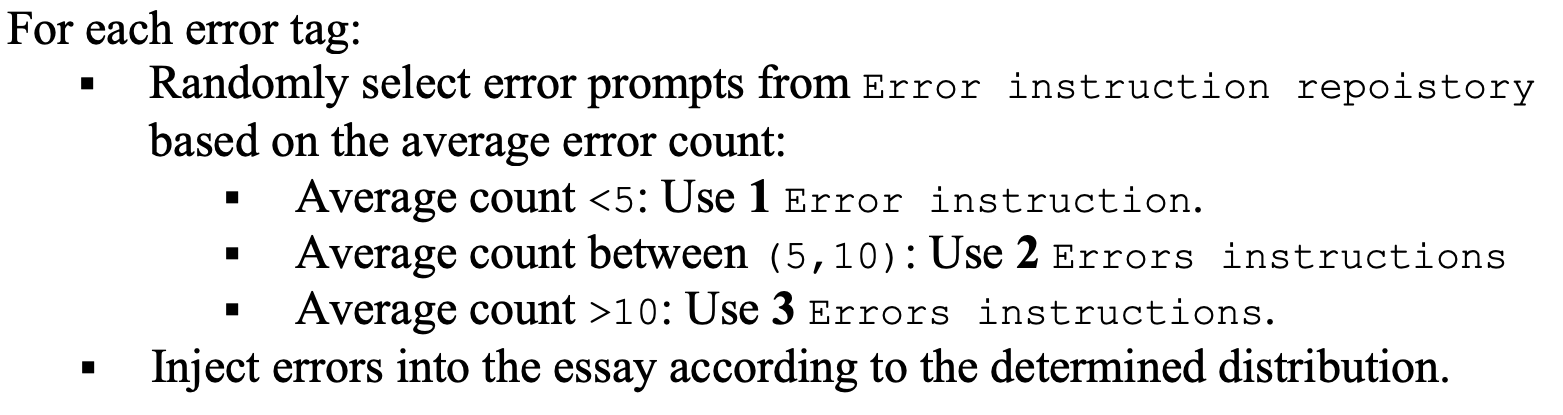}}  
  \caption{Error injection based on average error count}
  \label{fig:error_strategy}
\end{figure}

\subsection{Controlled Error Injection}

We introduce a controlled method for injecting errors into clean text, ensuring that the resulting erroneous sentences follow the empirical error distributions observed at each CEFR level. More formally, given an input sentence ($X$) and its CEFR level ($L$), we introduce errors in two steps: Error Type Prediction and Error Realization.

\paragraph{Error Type Prediction}
We estimate the probability of an error type occurring at a given word, i.e., $P(error\_type | word)$. To do so, we leverage ARETA in a reverse annotation process where we process correct--erroneous sentence pairs, tagging each correct word with its corresponding error type. Using this annotated data, we train a token-level BERT classifier to predict the most likely error type for each word in a given correct sentence. We fine-tune CAMeLBERT MSA \cite{inoue-etal-2021-interplay} to build our classifier.

\begin{figure*}[ht]
    \centering
    %\adjustbox{trim=10 10 10 10,clip}
    %{\includegraphics[width=0.8\linewidth]
    %\includegraphics[width=0.9\linewidth]
    %{Figures/essay_example_b1.pdf}}
    \includegraphics[width=0.8\linewidth]{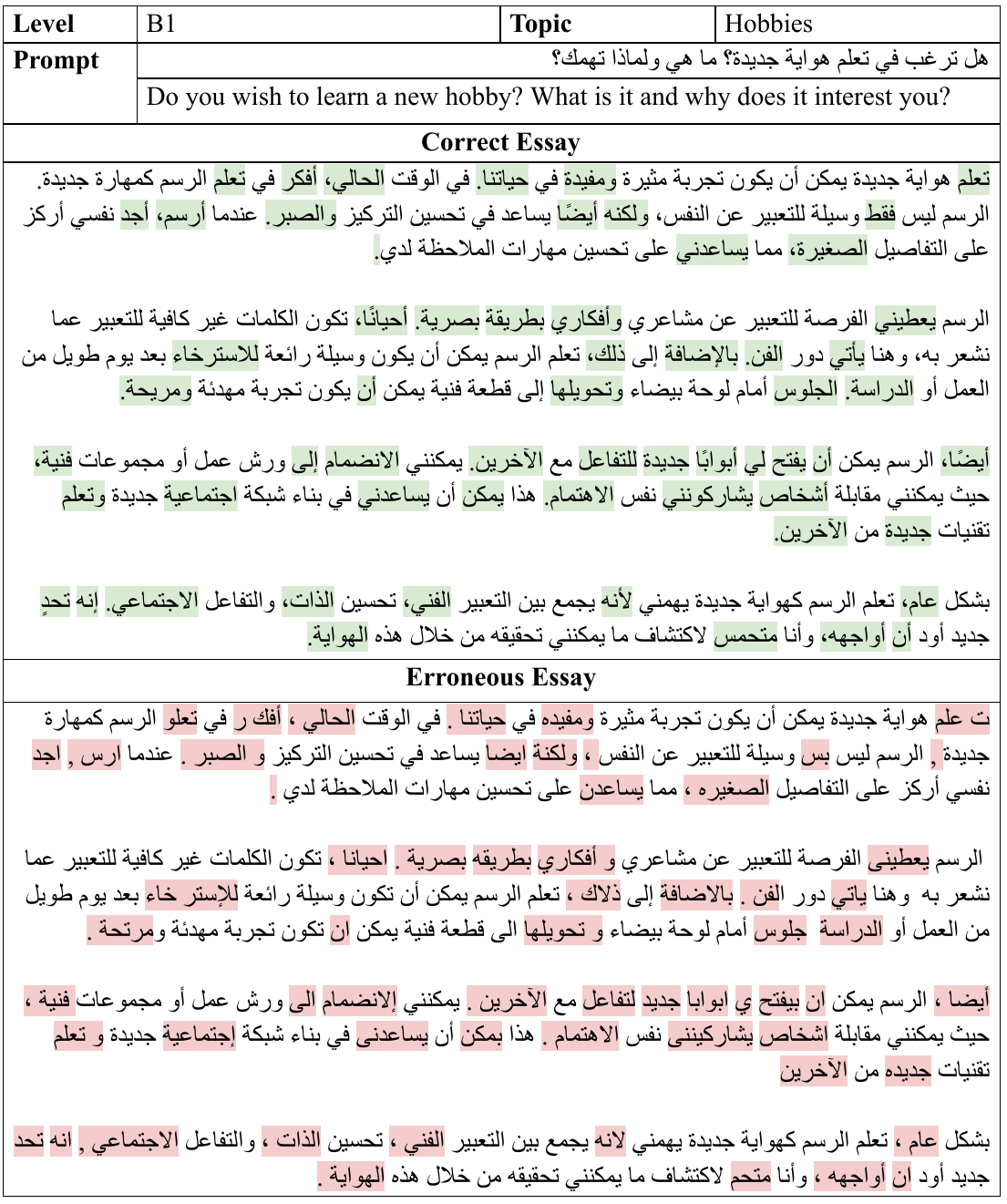}
    
    \caption{An Example of a B1 Arabic Essay generated by GPT-4o using the Hobbies prompt and the same essay after injecting errors by the controlled BERT-based model.}
    \label{fig:essay_pdf}
\end{figure*}

\paragraph{Error Realization}
To determine how a word should be corrupted, we first align correct--erroneous sentence pairs using the algorithm proposed by \newcite{alhafni-etal-2023-advancements}. For each aligned pair, we extract edit transformations that capture the operations required to convert a correct word into its erroneous counterpart. Using this data, we estimate $P(transformation | error\_type)$ with a bigram Maximum Likelihood Estimation (MLE) lookup model: $count(transformation, error\_type)$ / $count(error\_type)$.
During inference, we apply the BERT classifier to predict error types for each word in a sentence. We then filter these predictions, retaining only error types relevant to the sentence's CEFR level. Finally, the MLE model selects the most probable corruption for a given error type. A complete example of a B1--level essay generated by the proposed model is in 
Figure~\ref{fig:essay_pdf}.%Figure \ref{fig:essay_pdf} shows an example generated by the proposed method.

% %\input{Tables/essay_example}
% \begin{figure}[h]
% \centering 
% %\resizebox{\linewidth}{!}{
% \fbox{\includegraphics[width = 0.9\columnwidth]%[scale=0.5]
% {Figures/essay_example.png}
% }  
%  %\includegraphics[width=\columnwidth]{Figures/error_weight_approach.png}
%   \caption{An Example of a B1 Arabic Essay generated by GPT-4o using the Hobby prompt and the same essay after injecting errors by the controlled BERT-based model.}
%   \label{fig:essay_example}
% \end{figure}

% \begin{figure*}[ht]
%     \centering
%     %\adjustbox{trim=10 10 10 10,clip}
%     {\includegraphics[width=0.8\linewidth]
%     %\includegraphics[width=0.9\linewidth]
%     {Figures/essay_example_b1.pdf}}
%     \caption{An Example of a B1 Arabic Essay generated by GPT-4o using the Hobby prompt and the same essay after injecting errors by the controlled BERT-based model.}
%     \label{fig:essay_pdf}
% \end{figure*}
%\url{https://docs.google.com/document/d/1SdMLE3JD0TKGXImAUObHjQLIhJ2gOr9cWWERLtQe-xg/edit?usp=sharing}
%https://docs.google.com/document/d/1SdMLE3JD0TKGXImAUObHjQLIhJ2gOr9cWWERLtQe-xg/edit?usp=sharing

\begin{table*}[]
\resizebox{\linewidth}{!}{
    \centering
    \begin{tabular}{lccccc|ccccc}
        \toprule
          & \multicolumn{5}{c}{\textbf{Average Reference}} & \multicolumn{5}{|c}{\textbf{Multi-Reference}} \\
        %\textbf{Train Data} & \textbf{QWK (Avg-Ref)} & \textbf{Acc (Avg-Ref)} & \textbf{F1 (Avg-Ref)} & \textbf{P (Avg-Ref)} & \textbf{R (Avg-Ref)} & \textbf{QWK (Multi-Ref)} & \textbf{Acc (Multi-Ref)} & \textbf{F1 (Multi-Ref)} & \textbf{P (Multi-Ref)} & \textbf{R (Multi-Ref)} \\
        \textbf{Train Data} & \textbf{QWK} & \textbf{Acc} & \textbf{F\textsubscript{1}} & \textbf{P} & \textbf{R} &  \textbf{QWK} & \textbf{Acc} & \textbf{F\textsubscript{1}} & \textbf{P} & \textbf{R} \\
        \midrule
        ZAEBUC (baseline) & 22.44 & 57.58 & 24.50 & 23.33 & 26.76 & 61.06 & 84.85 & 43.70 & 42.50 & 45.31 \\
        
        ZAEBUC + GPT essays & 14.92 & 60.61 & 26.43 & 25.55 & 27.45 & 96.00 & 96.97 & 92.32 & 98.04 & 88.89 \\
        
        ZAEBUC + BERT errors & \textbf{27.87} & 57.58 & \textbf{38.02} & 35.86 & 44.93 & 82.70 & 87.88 & 71.66 & 70.83 & 74.38 \\
        
        ZAEBUC + GPT errors\_1 & 17.14 & 57.58 & 25.64 & 25.18 & 26.27 & \textbf{96.47} & 96.97 & 94.16 & 97.92 & 91.67 \\
        
        ZAEBUC + GPT errors\_0 & 20.84 & 57.58 & 32.76 & 31.53 & 46.08 & 93.79 & 93.94 & \textbf{95.12} & 96.49 & 94.44 \\
        \bottomrule
    \end{tabular}
   }
    \caption{Performance comparison of different training datasets. GPT essays are the original correct essays generated from GPT-4o, BERT errors are the erroneous essays using the controlled injection BERT model, GPT errors\_1 are the erroneous essays using GPT-4o with one-shot error example, while GPT errors\_0 with Zero-shot settings.}
    \label{tab:evaluation_results}
\end{table*}

\section{Experimental Setup}\label{evaluation}
This study focuses on introducing a data augmentation framework and synthetic Arabic essay corpus, rather than proposing a new AES model. We use a BERT-based model trained on the original ZAEBUC dataset as the reference baseline, evaluating how different augmentation strategies (e.g., GPT-4o generation, BERT-based error injection) improve performance relative to this setup.

\subsection{Data and Metrics} 
%We use Quadratic Weighted Kappa (QWK) \cite{Cohen:1968:weighted} as our primary evaluation metric, as it is the most widely-used metric in recent AES research \cite{ke:2019:automated}.
We use  the ZAEBUC dataset for all the experiments, following the splits created by \newcite{alhafni-etal-2023-advancements}: 70\% Train, 15\% Dev, and 15\% Test. 

Our primary evaluation metric is Quadratic Weighted Kappa (QWK) \cite{Cohen:1968:weighted},
the most widely used metric in AES research \cite{ke:2019:automated}.
We also report accuracy, macro precision (P), recall (R), and F\textsubscript{1} scores. Model predictions are evaluated in two settings settings: average-refence and multi-reference. The average-reference setting uses the rounded average of the three scores as the gold label, while the multi-reference considers each of the three human-assigned labels as a valid reference during evaluation,  following a more tolerant evaluation strategy (\S\ref{data}).
%the average of the gold holistic CEFR labels and in a multi-reference setting against the three individual gold labels per essay (\S\ref{data}).
%It is important to highlight that QWK offers a more reliable and appropriate metric for evaluating AES systems, particularly in the presence of imbalanced class distributions. Unlike accuracy, which can be biased toward dominant classes, QWK accounts for the ordinal nature of proficiency levels and penalizes predictions based on their distance from the true label. This makes it especially suited for assessing performance across a range of CEFR levels. As such, even modest gains in QWK indicate meaningful improvements in the model’s ability to distinguish between levels. These findings reinforce the value of using synthetically balanced data to reduce training bias and promote more equitable AES outcomes.

%We emphasize QWK as a superior metric for AES, particularly with imbalanced data. Unlike accuracy, QWK considers the ordinal nature of CEFR levels, penalizing predictions based on their distance from the true label. This makes it ideal for assessing performance across proficiency ranges, and even modest QWK improvements highlight the benefit of balanced synthetic data for fair AES.

%\subsection{Data} 
 %Table \ref{tab:evaluation_results} shows the performance results of the BERT-based Scoring Model utilizing different training sets. 

\subsection{Model}
% A BERT model is fine-tuned to evaluate Arabic essays based on CEFR levels.  (the evaluation aimed to gauge the models performance when exposed to a wide variety of data and diverse question types.)
% \begin{itemize}
%     \item various experiments based on Zeabuc as a training dataset.
%     \item Record the results for various testing and evaluation datasets (GPT data, rule-based data).
%     \textbf{(Bashar and Chatrine)}
%     \item final experiment to test how GPT functions as an auto-score.
% \end{itemize}
We treat the task of AES as a text classification problem. We fine-tune CAMeLBERT MSA \cite{inoue-etal-2021-interplay} on the training split of ZAEBUC. The models were trained by using the average CEFR gold labels. During training, we ignore the essays that are labeled as Unassessable, but  we penalize the models for missing them in the evaluation. We fine-tune the models for 5 epochs, with a maximum sequence length of 512, a learning rate of 5e-5, and a batch size of 32. 

\subsection{Results}
Our results are presented in Table~\ref{tab:evaluation_results}.
Our baseline system, only trained on the ZAEBUC training set, indicates room for improvement, with the F\textsubscript{1} at 24.50\% and  QWK at 22.44\%. We then switched between different datasets to %enhance the Arabic AES model and to 
measure the impact of data augmentation on the model. 
\paragraph{Impact of Synthetic Data} 
 We tested data augmentation by adding 3,040 corrected GPT-4o-generated essays, which lowered QWK but increased F\textsubscript{1}. Notably, the multi-reference setting saw significant gains, with QWK at 96.00\% and F\textsubscript{1} at 92.32\%.
This pattern stems from the flexibility of multi-reference evaluation, which treats all three human-assigned CEFR labels as valid references. This accommodates natural scoring variations and increases the chance that model predictions, especially on synthetic data, align with at least one reference label, boosting QWK and F\textsubscript{1} scores for both GPT-generated and error-injected essays.

\paragraph{Comparison of Error Injection Methods} As the initial synthetic essays were error-free, we further refined the model by adding essays with human-like errors. We compared two methods from \S\ref{error-injection}: (1) GPT-based error injection (with and without instruction examples) and (2) the controlled BERT-based method.

The results demonstrate that the controlled error model improves performance in all metrics, particularly in the average reference setting, which achieved 27.87 \% and 38.02 for QWK and F\textsubscript{1}, respectively.
This result aligns with expectations, as the BERT-injected errors closely follow CEFR-based error distributions, producing errors that realistically reflect learner writing and better match the average of human ratings.

GPT-based error injection performed best in the multi-reference setting, with one-shot examples reaching 96.47\% QWK and zero-shot boosting F\textsubscript{1} to 95.12\%. While less aligned with CEFR profiles, GPT errors benefit from fluency and variability, increasing the chance of matching at least one human reference in this flexible evaluation.

\section{Discussion} \label{discussion}
%This study demonstrated the effectiveness of using synthetic data and controlled error injection to enhance Arabic-AES. The results highlight several key insights regarding the interpretation of the metrics, the expansion of data and the methodological choices.
This study demonstrated the effectiveness of synthetic data and controlled error injection in enhancing Arabic AES, providing key insights into metric interpretation, data expansion, and methodological choices.

First, we emphasize that QWK offers a more robust metric than accuracy for evaluating AES systems, particularly under imbalanced class distributions. Unlike accuracy, which is biased by majority classes, QWK penalizes errors by their ordinal distance from the correct label. As Table 6 shows, even modest QWK improvements indicate meaningful advancements in differentiating CEFR levels, a distinction especially relevant given the skewed ZAEBUC dataset.
 %First, we emphasize that Quadratic Weighted Kappa (QWK) provides a more robust metric than accuracy for evaluating AES systems, particularly under imbalanced class distributions. While accuracy is susceptible to bias from majority classes, QWK penalizes errors according to their ordinal distance from the correct label. As shown in Table 6, even modest improvements in QWK suggest more meaningful advancements in the model's ability to differentiate between CEFR levels. This distinction is especially relevant given the skewed nature of the original ZAEBUC dataset.

The significant gains observed in the multi-reference setting  with generated GPT-4o essays stem from its flexibility. This evaluation approach treats all three human-assigned CEFR labels as valid references, accommodating natural scoring variations and increasing the chance that model predictions align with at least one reference label.

Our analysis revealed that while GPT-4o is powerful for generating diverse content, it struggles to precisely follow the nuanced distribution and specific linguistic features, including error patterns, observed in the manually annotated ZAEBUC dataset. In the GPT-based error injection approach, error type selection is guided by average error counts from the ZAEBUC corpus, but error realization depends on GPT-4o's interpretation of the prompt, making it less predictable. This inherent challenge in mimicking human-like linguistic and error distributions through zero-shot generation directly contributed to the observed lower agreement rate.

In contrast, the controlled method employs a BERT-based classifier for error prediction and applies transformations using bigram-MLE. This systematic approach resulted in a more robust replication of empirically observed error patterns, leading to its superior performance in the average-reference setting. This is expected, as BERT-injected errors more closely resemble learner writing and align more closely with average human ratings.

Overall, our findings highlight a trade-off between error alignment and fluency in data augmentation. Controlled error injection excels in the average-reference setting due to its closer alignment with learner errors, while GPT-based augmentation benefits from multi-reference flexibility but less reliably replicates authentic errors. The controlled BERT-based method thus serves as a key component of our pipeline, effectively addressing the limitations of direct GPT error injection.

\paragraph{Qualitative Analysis}
The qualitative analysis of the generation process revealed various biases in the GPT-4o outputs, including cultural, gender, and ideological biases. %For example, there were repeated references to traditional Arabic themes, stereotypical gender roles, and culturally narrow assumptions.
 For instance, the essays frequently referenced traditional Arabic themes, reinforced stereotypical gender roles, and reflected culturally narrow assumptions. A clear example of religious bias is that \<الجمعة> `Friday' was selected as \textit{the favorite day} in all 20 generated essays.  Additionally, there was a noticeable tendency to use masculine forms throughout the texts.  Such biases may unintentionally disadvantage students whose writing reflects different experiences, perspectives, or identities. Examples of these biases, along with their frequencies, are provided in Appendix \ref{secb: bias_example}.  We also observed a lack of diversity among the ten essays generated per prompt, with GPT-4o often repeating similar lexical and structural patterns. %Addressing these biases through broader training contexts and human validation will be essential in future work.

\section{Conclusions and Future Work}
\label{conclusions}
%This paper proposes a hybrid framework for an Arabic AES system, utilizing LLMs and transformers to address data scarcity by generating synthetic data that mimic Arabic learner writing.  Leveraging ZAEBUC corpus, we built linguistic and error profiles aligned with CEFR levels and prompted GPT-4o to generate 3,040 essays on various CEFR levels across 152 prompts. However, GPT-4o's effectiveness depends heavily on prompt engineering, achieving only 27.5\% alignment with reference profiles.
This paper presents a hybrid framework for Arabic AES, using LLMs and transformers to tackle data scarcity by generating synthetic essays that partly replicate Arabic learner writing. Building on the ZAEBUC corpus, we developed CEFR-aligned linguistic and error profiles and used GPT-4o to produce 3,040 essays across 152 prompts. However, GPT-4o’s performance relies heavily on prompt engineering, achieving only 27.5\% alignment with our reference profiles.

To introduce errors, %into GPT-generated essays
 we compare our two methods: (1) GPT-4o prompted multi-step error injection, and (2) our controlled method fine-tuning the CAMeLBERT MSA model to inject errors proportionally to their profiled occurrence.

Evaluated with a fine-tuned BERT classifier, our hybrid framework, combining GPT-generated data with controlled error injection, outperformed the baseline (QWK: 27.87\%, F\textsubscript{1}: 38.02\% ), offering more reliable and interpretable results. These findings demonstrate the effectiveness of controlled error injection in capturing learner error distributions across CEFR levels.
%% we can elaborate more for the FW on this part
%While our evaluation relied on automated methods, we acknowledge the importance of human assessment for fluency, naturalness, and CEFR alignment. Due to resource constraints, human evaluation was not feasible in this study; however, we plan to engage CEFR-trained raters in future work.

For future work, we will prioritize integrating a human evaluation into our framework. Human annotators will assess the fluency and naturalness of synthetic essays, as well as the realism of injected errors, ensuring that they reflect typical learner patterns at specific CEFR levels. %This validation is crucial for mitigating potential biases introduced by LLMs and enhancing trust in the generated data.

To improve generalizability, we also plan to expand the diversity of prompts beyond predefined topics %, enabling the AES system to handle a broader range of essay themes. In addition, our objective is to implement a broader 
and incorporate a wider set of writing traits, including coherence, logical flow, and topic relevance, beyond syntactic and lexical features.

%To strengthen the accuracy of our synthetic dataset, we will incorporate additional manually annotated essays, which, alongside ZAEBUC, will help capture the nuanced linguistic variations across CEFR levels. 
We also intend to enhance CEFR-level modelling by incorporating more manually annotated essays. This will help capture nuanced linguistic variations across levels and increase the robustness of our dataset.
%Lastly, we intend to conduct a detailed qualitative comparison between the GPT-injected and controlled error types to refine our error injection process and improve its realism.
%Looking ahead, we aim to expand the dataset by incorporating real essay samples and refining the linguistic and error profiles for each CEFR level to further enhance the data augmentation model. Additionally, we plan to deploy the AES system as an interactive educational writing assistant, providing students with immediate feedback on errors and CEFR scores to support Arabic writing development.
%Lastly, we aim to expand the dataset with real essays, refine CEFR-level profiles, and deploy the AES system as an interactive tool to provide students with instant feedback on errors and proficiency levels.
Lastly, we aim to deploy the AES system as an interactive tool to provide users with instant feedback on  errors and proficiency levels.

%\newpage

%\section*{Acknowledgments}

\section*{Limitations}

Despite the effectiveness of our hybrid Arabic AES framework, we note several limitations related to the quality of generated Arabic essays, error injection accuracy, and the generalization of the AES model.  The lack of A1 and C2 essays in ZAEBUC means that there is no gold reference data for these levels, which may impact both linguistic and error profiles, affecting the accuracy of GPT-generated essays. 
Furthermore, different biases are present in both the ZAEBUC dataset and GPT-4o outputs as discussed in (\S\ref{discussion})%. . ZAEBUC overrepresents  B-level essays, skewing model training. Meanwhile, GPT-4o exhibits cultural, gender, name-based, and ideological biases, which could disadvantage learners whose preferences or writing styles differ. %Addressing such bias remains an important direction for future research
%Furthermore, different biases exist in both ZAEBUC and the GPT outputs:ZAEBUC overrepresents B-level essays, skewing model training; and GPT-4o tends to favor Arabic culture themes, thereby limiting diversity in the generated texts and usability in other contexts. 
% For instance, when prompted to write about `favorite food,' most generated essays focused on traditional dishes like Kabsa \<كبسة> and Couscous \<كسكسي>, despite the fact that many students—especially younger ones—might prefer fast food or other regional dishes like Dolma \<دولمة>, Yalanji \<يلنجي>, or Mahshi \<محشي>. 
% None of the generated essays explored these alternative food choices.This pattern suggests that the model may favor certain cultural norms, which could disadvantage learners whose preferences or writing styles differ. Addressing such cultural bias remains an important direction for future research.

In addition, due to the lack of comprehensive gold data, GPT struggles to fully replicate real learner writing styles, achieving only 27.5\% agreement with linguistic feature profiles.

Another limitation is the model’s ability to generalize across various domains and question types. The AES system may struggle with broader writing tasks and alternative prompts since the dataset and augmentation methods focus on predefined prompts. Relying solely on CEFR as a holistic scoring method limits interpretability. Enhancing the dataset with multi-trait annotations, such as coherence, argumentation, and organization, could improve scoring accuracy and feedback quality. Moreover, better-controlled GPT prompting could refine the quality and diversity of generated essays, reducing biases and improving alignment with real learner writing patterns.

%While our evaluation relied on automated methods, we acknowledge the importance of human assessment for fluency, naturalness, and CEFR alignment. 
Due to resource constraints, human evaluation was not feasible in this study; however, we plan to engage CEFR-trained annotators in the future.

\section*{Ethical Considerations}
While Arabic AES systems provide significant support in assessing Arabic learners' writing proficiency, it is essential to highlight the ethical implications of their use. Automatic assessment and scoring may lead to misjudgments that could distress learners and students, especially if their work is incorrectly evaluated. AES tools should serve as an educational assistive technology, complementing the teacher's judgment, not replacing it in educational settings.

% Bibliography entries for the entire Anthology, followed by custom entries
%\bibliography{anthology,custom}
% Custom bibliography entries only

%\newpage
\bibliography{custom,anthology}

\onecolumn
\appendix
%\appendix
%\newpage
\section{GPT-4o Generated Essays}
\subsection{Statistics}
\label{secb:appendix}
\begin{table}[h]
\resizebox{\linewidth}{!}{

\begin{tabular}{|c|c|c|c|c|c|c|c|c|}
\hline
\multicolumn{1}{|l|}{\textbf{CEFR}} &
  \multicolumn{1}{l|}{\textbf{\#Essays}} &
  \multicolumn{1}{l|}{\textbf{\#Words}} &
  \multicolumn{1}{l|}{\textbf{\#Sentences}} &
  \multicolumn{1}{l|}{\textbf{\#Tokens}} &
  \multicolumn{1}{l|}{\textbf{Avg\_W\_L}} &
  \multicolumn{1}{l|}{\textbf{Avg\_S\_L}} &
  \multicolumn{1}{l|}{\textbf{Unique\_Tokens}} &
  \multicolumn{1}{l|}{\textbf{Unique\_Words}} \\ \hline
\textbf{A1}                          & 470  & 39367  & 6721  & 47612  & 4.66 & 6.08  & 4518  & 4512  \\ \hline
\textbf{A2}                          & 470  & 54980  & 5565  & 63769  & 4.82 & 10.46 & 7147  & 7143  \\ \hline
\textbf{B1}                          & 530  & 100185 & 7276  & 113672 & 4.97 & 14.62 & 10742 & 10734 \\ \hline
\textbf{B2}                          & 530  & 127029 & 7804  & 142995 & 5.03 & 17.32 & 12522 & 12509 \\ \hline
\textbf{C1}                          & 520  & 136849 & 7630  & 152962 & 5.16 & 19.05 & 12832 & 12823 \\ \hline
\textbf{C2}                          & 520  & 146253 & 7946  & 163461 & 5.16 & 19.57 & 13686 & 13676 \\ \hline
\multicolumn{1}{|l|}{\textbf{Total}} & \textbf{3040} & \textbf{604663} & \textbf{42942} & \textbf{684471} &\textbf{ 5.04} & \textbf{14.94} & \textbf{27286} & \textbf{27272} \\ \hline
\end{tabular}
}
\caption{Summary statistics of the generated Arabic synthetic essay corpus across CEFR levels. \#Essays denotes the number of essays; \#Words refers to the total word count; \#Sentences indicates the total number of sentences; \#Tokens represents the total number of tokens (vocabulary items); Avg\_W\_L corresponds to the average word length in characters; Avg\_S\_L refers to the average sentence length in words; lexical diversity is captured through the counts of unique tokens and unique words.}
\label{tab:ARWI_statistics}

\end{table}

\subsection{Bias in the Generated Essays}
\label{secb: bias_example}
\begin{table}[h]
\resizebox{\linewidth}{!}{
\begin{tabular}{|r|l|c|c|c|l|}
\hline
\multicolumn{1}{|c|}{\textbf{Arabic Prompt}} &
  \multicolumn{1}{c|}{\textbf{English Prompt}} &
  \multicolumn{1}{c|}{\textbf{Arabic Response}} &
  \multicolumn{1}{c|}{\textbf{English Response}} &
  \textbf{Occurrences /20} &
  \multicolumn{1}{c|}{\textbf{Bias}} \\ \hline 
  \< ما هو يومك المفضل؟              > & What is your favorite day?                  &
  \<الجمعة         >  & Friday              & 20 & Cultural/Religious Bias    \\ \hline
  \<
ماهو طعامك المفضل؟            >   & What is your favorite food?                 & \< البيتزا          >& Pizza               & 16 & Globalization Bias         \\ \hline
\<
ما هي هوايتك المفضلة؟          >  & What is your favorite hobby?                &\< القراءة        >  & Reading             & 10 & Socioeconomic/Class Bias   \\ \hline
\<
رحلة ذهبت إليها               >   & A trip you went on                          & \< ذهبت إلى البحر  > & I went to the beach & 18 & Geographical/Cultural Bias \\ \hline
\<
ماذا تفعل في عطلة نهاية الأسبوع؟ >& What do you do on the weekend?              &\< نذهب إلى الحديقة >& We go to the park   & 20 & Geographical/Cultural Bias \\ \hline
\<
ماهي المادة الدراسية المفضلة؟   > & What is your favorite school subject?       & \< الرياضيات  >      & Mathematics         & 17 & Educational System Bias    \\ \hline
\<
ما هي رياضتك المفضلة؟         >   & What is your favorite sport?                &\< كرة القدم    >    & Football            & 20 & Cultural Bias                \\ \hline
\<
شخص تعتبره مثلك الأعلى        >   & A person you consider your role model       &\< والدي       >     & My father           & 17 & Gender Bias                \\ \hline
\<
من هو أفضل صديق لك؟     >         & Who is your best friend?                    &\< أحمد     >        & Ahmed               & 20 & Gender/Name Bias           \\ \hline
\<
المهنة المستقبلية    >            & Your future profession                      &\< طبيب     >        & Doctor              & 13 & Stereotype Bias            \\ \hline
\<
معلم شهير في العالم الأدبي   >    & A famous teacher in the world of literature & \< الأهرامات    >    & The Pyramids        & 17 & Cultural Bias              \\ \hline
\<
أفضل فصول السنة            >      & Your favorite season of the year            &\< الصيف     >       & Summer              & 20 & Climate Bias               \\ \hline
\<
لغة تود تعلمها         >          & A language you would like to learn          & \<الأسبانية  >      & Spanish             & 16 & Language Bias              \\ \hline
\<
بلد ترغب في السفر إليه    >       & A country you would like to visit           &\< مصر   >           & Egypt               & 10 & National Identity Bias     \\ \hline
\<
قوة خارقة تتمناها        >        & A superpower you wish to have               &\< الطيران     >     & Flying              & 19 & Media Bias                 \\ \hline
\end{tabular}
}
\caption{Examples of response biases in GPT-4o generated essays. }
\label{tab:bias-examples}
\end{table}
\newpage

\section{GPT-4o Error Injection Algorithm}
\label{sec:appendix}
\begin{figure}[h]
\centering 
%\resizebox{\linewidth}{!}{
\fbox{\includegraphics[width=0.80\linewidth]%[scale=0.5]
{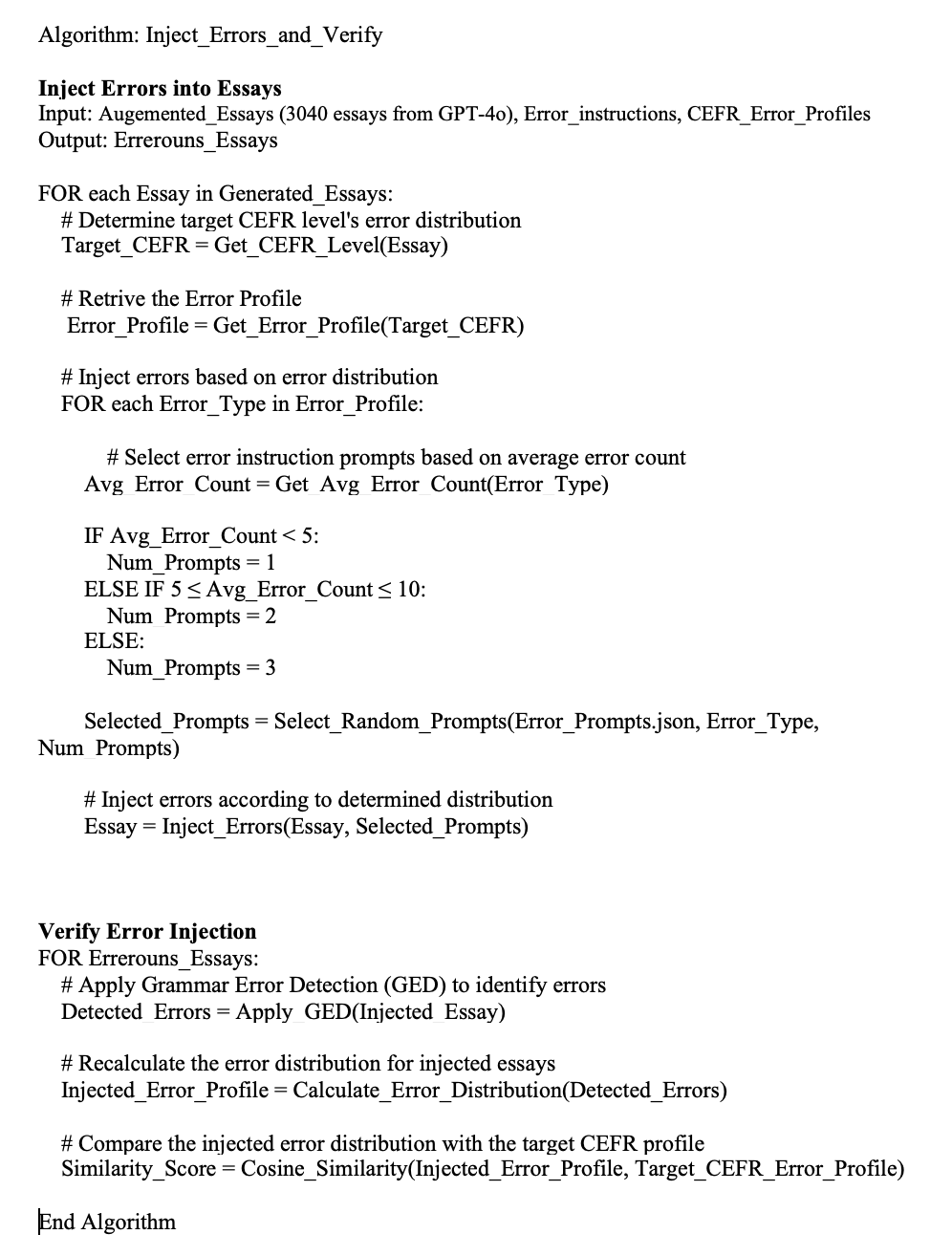}
}  
 %\includegraphics[width=\columnwidth]{Figures/error_weight_approach.png}
% \caption{GPT-4o Error injection and verification algorithm}
  \label{fig:error_injection}
\end{figure}

\newpage
%\vspace*{3cm}
\section{Error Types Taxonomy}
\label{sec:error_taxonomy}
\begin{table}[h]
\resizebox{0.95\linewidth}{!}{
\begin{tabular}{c|c|l|l|c|c|}
\cline{2-6}
 &
  \textbf{11-Classes} &
  \multicolumn{1}{c|}{\textbf{42-Classes}} &
  \multicolumn{1}{c|}{\textbf{Error Description}} &
  \textbf{Correct  Word} &
  \textbf{Erroneous Word} \\ \hline
\multicolumn{1}{|c|}{\multirow{2}{*}{\begin{tabular}[c]{@{}c@{}}Morphology\\ (M)\end{tabular}}} &
  \multirow{2}{*}{M} &
  MI &
  Inflection &
 \< عارف> &
\<  معروف> \\ \cline{3-6} 
\multicolumn{1}{|c|}{} &
   &
  MT &
  Tense &
 \< ذهب> &
 \< يذهب> \\ \hline
\multicolumn{1}{|c|}{\multirow{18}{*}{\begin{tabular}[c]{@{}c@{}}Orthography\\ (O)\end{tabular}}} &
  \multirow{18}{*}{O} &
  OA &
  Alef-Maqsura &
 \< القاضي> &
 \< القاضى >\\ \cline{3-6} 
\multicolumn{1}{|c|}{} &
   &
  OA+OH &
  Alef-Maqsura + Hamza &
 \< أضحى >&
 \< اضحا >\\ \cline{3-6} 
\multicolumn{1}{|c|}{} &
   &
  OA+OR &
  Alef-Maqsura + Wrong Character &
\<  كشيء> &
 \< كشىء> \\ \cline{3-6} 
\multicolumn{1}{|c|}{} &
   &
  OC &
  Chatacter Order &
 \< المدرسة >&
\<  المردسة> \\ \cline{3-6} 
\multicolumn{1}{|c|}{} &
   &
  OD &
  Extra Character &
 \< هذا> &
 \< هاذا> \\ \cline{3-6} 
\multicolumn{1}{|c|}{} &
   &
  OD+OG &
  Extra Character + Lengthening Short Vowels &
 \< تتطورو >&
 \< تتطور >\\ \cline{3-6} 
\multicolumn{1}{|c|}{} &
   &
  OD+OH &
  Extra Character + Hamza &
 \< لأنهم >&
 \< ألانهم >\\ \cline{3-6} 
\multicolumn{1}{|c|}{} &
   &
  OD+OM &
  Extra Character + Missing Character &
\<  الاجتماعي >&
\<  الاجتاعيي >\\ \cline{3-6} 
\multicolumn{1}{|c|}{} &
   &
  OD+OR &
  Extra Character + Wrong Character &
\<  الصور >&
\<  السوور> \\ \cline{3-6} 
\multicolumn{1}{|c|}{} &
   &
  OH &
  Hamza &
\<  العب >&
 \< إلعب >\\ \cline{3-6} 
\multicolumn{1}{|c|}{} &
   &
  OH+OM &
  Hamza + Missing Character &
 \< الأشياء >&
\<  الاشيا > \\ \cline{3-6} 
\multicolumn{1}{|c|}{} &
   &
  OH+OT &
  Hamza + Ta-Marbuta &
\<  إمارة >&
 \< اماره > \\ \cline{3-6} 
\multicolumn{1}{|c|}{} &
   &
  OM &
  Missing Character &
\<  المدرسة >&
\<  المدسة >\\ \cline{3-6} 
\multicolumn{1}{|c|}{} &
   &
  OM+OR &
  Missing Character + Wrong Character &
\<  المجتمع >&
\<  الجطمع >\\ \cline{3-6} 
\multicolumn{1}{|c|}{} &
   &
  OR &
  Wrong Character &
 \< المدرسة> &
 \< المدرصة >\\ \cline{3-6} 
\multicolumn{1}{|c|}{} &
   &
  OR+OT &
  Wrong Character + Ta-Marbuta &
  \<مكتظة> &
 \< مكتضه> \\ \cline{3-6} 
\multicolumn{1}{|c|}{} &
   &
  OT &
  Ta-Marbuta &
\<  غرفة> &
 \< غرفه >\\ \cline{3-6} 
\multicolumn{1}{|c|}{} &
   &
  OW &
  Alef-Fariqa &
\<  كتبوا >&
 \< كتبو >\\ \hline
\multicolumn{1}{|c|}{\multirow{2}{*}{\begin{tabular}[c]{@{}c@{}}Semantics\\ (S)\end{tabular}}} &
  \multirow{2}{*}{S} &
  SF &
  Conjunction &
\<  فسبحان >&
\<  سبحان >\\ \cline{3-6} 
\multicolumn{1}{|c|}{} &
   &
  SW &
  Word Selection &
\<  على >&
\<  من >\\ \hline
\multicolumn{1}{|c|}{\begin{tabular}[c]{@{}c@{}}Punctuation\\ (P)\end{tabular}} &
  P &
  P &
  Punctuation &
  \multicolumn{1}{ c|}{\<السوق،>} &
  \multicolumn{1}{c|}{\<السوق.>} \\ \hline
\multicolumn{1}{|c|}{\multirow{8}{*}{\begin{tabular}[c]{@{}c@{}}Syntax\\ (X)\end{tabular}}} &
  \multirow{8}{*}{X} &
  XC &
  Case &
  \multicolumn{1}{c|}{\<رائعا>} &
  \multicolumn{1}{c|}{\<رائع>} \\ \cline{3-6} 
\multicolumn{1}{|c|}{} &
   &
  XC+XG &
  Case + Gender &
  \multicolumn{1}{c|}{\<مجتهدا>} &
  \multicolumn{1}{c|}{\<مجتهدة>} \\ \cline{3-6} 
\multicolumn{1}{|c|}{} &
   &
  XC+XN &
  Case + Number &
  \multicolumn{1}{c|}{\<نواح>} &
  \multicolumn{1}{c|}{\<نواحي>} \\ \cline{3-6} 
\multicolumn{1}{|c|}{} &
   &
  XF &
  Definiteness &
  \multicolumn{1}{c|}{\<المفيد>} &
  \multicolumn{1}{c|}{\<مفيد>} \\ \cline{3-6} 
\multicolumn{1}{|c|}{} &
   &
  XG &
  Gender &
  \multicolumn{1}{c|}{\<كان>} &
  \multicolumn{1}{c|}{\<كانت>} \\ \cline{3-6} 
\multicolumn{1}{|c|}{} &
   &
  XM &
  Missing Word &
  \multicolumn{1}{c|}{\<على>} &
  \multicolumn{1}{c|}{NULL} \\ \cline{3-6} 
\multicolumn{1}{|c|}{} &
   &
  XN &
  Number &
  \multicolumn{1}{c|}{\<كتابين>} &
  \multicolumn{1}{c|}{\<كتب>} \\ \cline{3-6} 
\multicolumn{1}{|c|}{} &
   &
  XT &
  Unnecessary Word &
  \multicolumn{1}{c|}{NULL} &
  \multicolumn{1}{c|}{\<على>} \\ \hline
\multicolumn{1}{|c|}{\multirow{2}{*}{\begin{tabular}[c]{@{}c@{}}Combination\\ (Comb.)\end{tabular}}} &
  M+O &
  MI+OH &
  Inflection + Hamza &
 \<  أشخاص >&
 \< اشخاصك >\\ \cline{2-6} 
\multicolumn{1}{|c|}{} &
  O+X &
  OH+XC &
  Hamza + Case &
 \< أضرارا> &
 \< اضرار >\\ \hline
\multicolumn{1}{|c|}{SPLIT} &
  SPLIT &
  SPLIT &
  Split &
  \multicolumn{1}{c|}{\<دولة  الإمارات>} &
  \multicolumn{1}{c|}{\<دولةالإمارات>} \\ \hline
  \multicolumn{1}{|c|}{MERGE} &
  MERGE &
  MERGE &
  Merge &
  \< بالعلم > &
\< ب العلم >\\ \hline
% \multicolumn{1}{|c|}{\multirow{2}{*}{MERGE}} &
% MERGE-B &
%   MERGE-B &
%   Merge &
%  \< بالعلم >&
%  \< ب العلم> \\ \cline{2-6} 
% \multicolumn{1}{|l|}{} &
% MERGE-I
%    &
%   MERGE-I &
%   Merge &
%  \< بالعلم >&
% \<  ب العلم >\\ \hline
\multicolumn{1}{|c|}{DELETE} &
  \multicolumn{1}{c|}{DELETE} &
  DELETE &
  \multicolumn{1}{l|}{Delete} &
  NULL &
  \multicolumn{1}{c|}{\<داخل>} \\ \hline
\end{tabular}
}
\caption{Illustrative examples of error types categorized according to the ARETA error taxonomy \cite{belkebir-habash-2021-automatic}. The table presents hierarchical mappings from coarse-grained (11-Class) to fine-grained (42-Class) error categories, alongside representative corrections. }

\label{tab:error_taxonomy}
\end{table}

\begin{comment}
   
 \newpage
 \section{Example from our Arabic Essay dataset}
 \label{sec:appendix_b}
\begin{figure}[ht]
    \centering
    %\adjustbox{trim=10 10 10 10,clip}
    %{\includegraphics[width=0.8\linewidth]
    %\includegraphics[width=0.9\linewidth]
    %{Figures/essay_example_b1.pdf}}
    \includegraphics[width=0.8\linewidth]{Figures/essay_example_b1_2.pdf}
    
    \caption{An Example of a B1 Arabic Essay generated by GPT-4o using the Hobby prompt and the same essay after injecting errors by the controlled BERT-based model.}
    \label{fig:essay_pdf}
\end{figure}
%\url{https://docs.google.com/document/d/1SdMLE3JD0TKGXImAUObHjQLIhJ2gOr9cWWERLtQe-xg/edit?usp=sharing}
%https://docs.google.com/document/d/1SdMLE3JD0TKGXImAUObHjQLIhJ2gOr9cWWERLtQe-xg/edit?usp=sharing
 
\end{comment}

\end{document}